%% file: main_document.tex
\documentclass[review]{elsarticle}
\usepackage{supertabular}
\usepackage{lineno,hyperref}
\modulolinenumbers[5]

\hypersetup{
    colorlinks = true,
    linkcolor = blue,
    citecolor = blue,
    urlcolor  = blue
}
\usepackage{caption}
\usepackage{amsmath}
\usepackage{amssymb}
\usepackage{epstopdf}

\usepackage{graphicx}
\usepackage{algorithm}
\usepackage{algpseudocode}
\usepackage{bm}
\usepackage{xcolor}
\usepackage{longtable}
\usepackage{framed} 
\usepackage{multicol} 
\usepackage{tabu}

\definecolor{review1}{rgb}{0,0,0} 

\definecolor{review2}{rgb}{0,0,0} 

\definecolor{red}{rgb}{1,0,0}
\definecolor{blue}{rgb}{0,0,1}

\begin{document}

\begin{frontmatter}

\title{Probabilistic Energy Forecasting using Quantile Regressions based on a new Nearest Neighbors Quantile Filter}



\author[myaddress]{Jorge \'Angel Gonz\'alez Ordiano\corref{mycorrespondingauthor}}
\ead{jorge.ordiano@kit.edu}
\author[myaddress]{Lutz Gr\"oll}
\author[myaddress]{Ralf Mikut}
\author[myaddress]{Veit Hagenmeyer}
\address[myaddress]{Institute for Automation and Applied Informatics, \\ Karlsruhe Institute of Technology, \\ Hermann-von-Helmholtz-Platz 1 \\ 76344 Eggenstein-Leopoldshafen}


\begin{abstract}
Parametric quantile regressions are a useful tool for creating probabilistic \textcolor{review1}{energy} forecasts. Nonetheless, since classical quantile regressions are trained using a non-differentiable cost function, their creation using complex data mining techniques (e.g., artificial neural networks) may be complicated. This article presents a method that uses a new nearest neighbors quantile filter to obtain quantile regressions independently of the utilized data mining technique and without the non-differentiable cost function. Thereafter, a validation of the presented method using the dataset of the Global Energy Forecasting Competition of 2014 is undertaken. 
\textcolor{review1}{The results show that the presented method is able to solve the competition's task with a similar accuracy and in a similar time as the competition's winner, but requiring a much less powerful computer. This property may be relevant in an online forecasting service for which the fast computation of probabilistic forecasts using not so powerful machines is required.}
\end{abstract}


\end{frontmatter}

\linenumbers

\footnotesize 
\textcolor{review1}{\input{nomenclature.tex}}
\normalsize
\nolinenumbers

\section{Introduction}\label{sec. introduction}

\textcolor{review1}{Integrating volatile renewable power systems (e.g., wind and photovoltaic (PV) power plants) into the electrical grid has complicated the necessary balancing of electricity demand and supply~\cite{Gottwalt17,Ludwig17}. Therefore, forecasts have become necessary to correctly plan and schedule the electrical grid~\cite{Dannecker15}.}

Time series forecasting models are a tool used for estimating the future development of values whose change over time is of interest (e.g., renewable generation and load)~\cite{Hyndman14}. Most forecasting models described in literature can be classified as point forecasting models, i.e. they deliver a single value at a given forecast horizon~\cite{Gneiting11}, yet they are unable to quantify their own forecast uncertainty. Probabilistic forecasting models, on the contrary, are able to quantify such uncertainty by delivering, for example, intervals with a given probability of a future value laying within, or probability distribution functions of a forecast time series value~\cite{Gneiting14}. Probabilistic forecasts have become an important decision making tool~\cite{Gneiting14}, since quantifying the forecast uncertainty may lead to better decisions. The current development of the ``Smart Grid"~\cite{Fang12} has found probabilistic forecasts both useful and necessary, as they are able to describe the inherent uncertainty of future renewable power generation and load (i.e. energy demand)~\cite{Hong16}. 

Probabilistic forecasting models can mostly be divided into parametric and non-parametric~\cite{GonzalezOrdiano18}. \textcolor{review1}{While the former assume that the uncertainty will follow a given probability distribution (e.g., Gaussian), the latter do not. In the context of energy forecasting, quantile regressions are one of the most commonly used approaches for obtaining non-parametric probabilistic forecasts  -- especially in wind power forecasting. Quantile regressions are able to estimate quantiles of a future time series value conditioned on the regressions' input~\cite{Fahrmeir13}, hence their combination allows for the creation of probabilistic forecasts, as e.g., interval forecasts. Some examples of quantile regressions being used for wind power forecasting are given by Brenmes~\cite{Bremnes04}, Haque et al.~\cite{Haque14}, and Nielsen et al.~\cite{Nielsen06}. The first utilizes local linear quantile regressions, while the others create quantile regressions based on linear combinations of basis functions. In the case of load, Gaillard et al.~\cite{Gaillard16} create probabilistic forecasts using linear quantile regressions with non-linear functions of the used features as covariates, while Liu et al.~\cite{Liu17} train linear quantile regressions using the forecasts of several point forecasting models as independent variables. Similarly in the case of PV power, quantile regressions have also been found useful. For example, Nagy et al.~\cite{Nagy16} use a combination of a quantile regression forest and a stacked random forest to estimate the forecast uncertainty. It is important to mention, that quantile regressions are not the only approach for describing the forecast uncertainty. Other techniques are shown for example, in the works of Zhang and Wang~\cite{Zhang15}, Juban et al.~\cite{Juban07}, and Xie and Hong~\cite{Xie16}. In the first work, PV power is forecast using a traditional k-nearest neighbors regression and a kernel density estimator, in the second  wind power probabilistic forecasts are obtained again with a kernel density estimator, and in the third a scenario-based probabilistic forecast together with a postprocessing step is used to obtain probabilistic load forecasts. A more in-depth description of these and other methods is out of the scope of the present contribution. Therefore, interested readers are referred to survey articles presented in~\cite{Antonanzas16}, \cite{Hong16a}, and~\cite{Zhang14} for more information regarding PV power, load, and wind power probabilistic forecasting respectively.}

Classical parametric quantile regressions are obtained by minimizing the non-differentiable sum of pinball-losses, a procedure that increases the difficulty of creating quantile regressions with more complex data mining techniques (e.g., ANNs and support vector regressions (SVRs)). One of the reasons thereof is that the lack of differentiability may lead first, to problems when using training algorithms based on gradient based optimization~\cite{Cannon11} and second, to higher \textcolor{review1}{computation times}. \textcolor{review1}{Furthermore, minimizing the sum of pinball-losses makes the utilization of ``out of the box" regression training algorithms (i.e. already implemented algorithms found in typical statistic/machine learning libraries) impossible, since they normally minimize other cost functions. Therefore, training classical parametric quantile regression requires the additional effort of modifying the ``out of the box" training algorithm (as shown in~\cite{Cannon11},~\cite{Shim16}, and~\cite{Taylor00}) or of programming a new training algorithm from scratch.}


The present contribution offers a generic approach that simplifies and allows the creation of linear \textcolor{review1}{and non-linear} parametric quantile regressions. The \textcolor{review1}{presented} approach expands a preliminary concept~\cite{GonzalezOrdiano16a} that has been applied for renewable energy scheduling~\cite{Appino18}. The new method is based on a newly developed nearest neighbors quantile filter (NNQF) that modifies the used training set. \textcolor{review1}{By doing so, the need of minimizing the sum of pinball-losses is eliminated, hence allowing the creation of parametric quantile regression using any regression data mining technique and their ``out of the box" training algorithms. In the present work, the developed method is validated using the data from the solar track of the Global Energy Forecasting Competition of 2014 (GEFCom14) as benchmark~\cite{Hong16}.}


Additionally, the obtainment of parametric quantile regressions without minimizing the sum of pinball-losses is further motivated by the desire of keeping the computational effort for both their training and application as low as possible. The reason thereof is to allow the deployment of these quantile regressions as part of an online forecasting service, which may require a fast computation of probabilistic forecasts. \textcolor{review1}{For instance,} the service planned for the Energy Lab 2.0~\cite{Duepmeier15,Hagenmeyer16}. 

The present contribution is organized as follows: Section~\ref{sec. state_of_the_art} briefly presents \textcolor{review1}{background information} of both classical parametric quantile regressions and time series forecasting. Section~\ref{sec. quant_reg_w_knn_training_set} gives a description of the new presented method, followed by Section~\ref{sec. benchmark} describing the conducted validation procedure. Section~\ref{sec. results} shows the obtained results and finally, Section~\ref{sec. conclusion} offers the conclusion and outlook of this work. 


\section{Background Information}\label{sec. state_of_the_art}


\subsection*{Classical Parametric Quantile Regressions}

\textcolor{review1}{Regression can be defined as a supervised learning approach that uses data mining techniques to obtain a data-driven model able to estimate an output value $y$ given an input vector $\mathbf{x}$~\cite{Fayyad96}. Parametric regression models, which are the ones relevant in the present contribution, can be defined as
\begin{equation}\label{eq. parametric_regression_with_estparam}
\hat{y} = f(\mathbf{x};\hat{\boldsymbol{\theta}}) \text{ .}
\end{equation}
with $f(\cdot)$ representing the regression model, $\hat{y}$ the estimate of the desired output, and $\hat{\boldsymbol{\theta}}$ the estimated regression parameters. The values in $\hat{\boldsymbol{\theta}}$ are estimated by minimizing a cost function on a training set comprised of $N$ different observations~\cite{Hastie08,Stulp15}. These observations are contained in an input matrix $\mathbf{X}$ and in a desired output vector $\mathbf{y}$ defined as:
\begin{equation}\label{eq. training_set_regression}
\begin{aligned}
\mathbf{X}     & = [\mathbf{x}_{1}, \cdots, \mathbf{x}_{n}, \cdots, \mathbf{x}_{N}]^{T} \\
\mathbf{y} & = [y_{1}, \cdots, y_{n}, \cdots,y_{N}]^{T}\text{ ;} 
\end{aligned}
\end{equation}}

A commonly used cost function (for example, in the case of a linear regression) is the sum of squared errors, whose minimization results in $\hat{y}$ being an estimate of the conditional expected value of $y$ given an input $\mathbf{x}$~\cite{Fahrmeir13}. Quantile regressions, on the contrary, estimate the conditional $q$-quantile of $y$ -- with $q\in(0,1)$ -- instead of its conditional expected value~\cite{Fahrmeir13,Koenker05}. In other words, a quantile regression is able to obtain an estimate of a value $y_{(q)}$ whose probability of being greater than or equal to $y$ given an input $\mathbf{x}$ is equal to $q$. In the present contribution, parametric quantile regressions are described as
\begin{equation}\label{eq. parametric_quantile_regression}
 \hat{y}_{(q)} = f(\mathbf{x};\hat{\boldsymbol{\theta}}_{(q)}) \text{ ;}
\end{equation}
with $\hat{y}_{(q)}$ being the estimate of the conditional $q$-quantile of $y$ and $\hat{\boldsymbol{\theta}}_{(q)}$ being the estimated parameters. In case of a classical quantile regression, $\hat{\boldsymbol{\theta}}_{(q)}$ is obtained by minimizing the non-differentiable sum of pinball-losses~\cite{Juban16}, i.e.
\begin{equation}\label{eq. pinball_loss_opt}
\hat{\boldsymbol{\theta}}_{(q)} = \underset{\boldsymbol{\theta}_{(q)}}{\operatorname{argmin}} \sum_{n=1}^{N}
\begin{cases}
(q-1)\cdot(y_{n}-f(\mathbf{x}_{n};\boldsymbol{\theta}_{(q)})) & \text{, if } y_{n} < f(\mathbf{x}_{n};\boldsymbol{\theta}_{(q)})\\
q\cdot(y_{n}-f(\mathbf{x}_{n};\boldsymbol{\theta}_{(q)})) & \text{, else} \text{ .}
\end{cases}
\end{equation}
\textcolor{review1}{Equation~\eqref{eq. pinball_loss_opt} may complicate the creation of classical quantile regressions with more complex data mining techniques, since:
\begin{itemize}
\item Minimizing the non-differentiable cost function may lead to problems with gradient based optimization (i.e. the approach commonly used for training neural networks)
\item The out of the box training algorithms (i.e. the ones found in existing machine learning libraries) have to be modified for them to solve Equation~\eqref{eq. pinball_loss_opt} or some new training algorithms have to be programmed from scratch.
\end{itemize}
Therefore, the present contribution offers an alternative.}

\subsection*{Time Series Forecasting}

The goal of a time series forecasting model is to estimate the future development of a time series at a given forecast horizon $H$, using available information. A finite time series $\{\mathsf{y}[k]; k = 1, \dots, K \}$ can be defined as a set of observations $\mathsf{y}[k]$ that are measured at specific equidistant points in time~\cite{Brockwell06}; such observations form a sequence in which the position of each observation is determined by their corresponding timestep value $k$. A multivariate finite time series $\{\bm{\mathsf{u}}[k]; k = 1, \dots, K \}$ can be defined in a similar manner, with the only difference being that its observations $\bm{\mathsf{u}}[k]$ are vectors formed by observations of various time series. 


The present contribution generalizes time series forecasting models as regressions, based on the fact that data mining techniques (as e.g., linear regressions, ANNs, and SVRs) have been found useful at forecasting future time series values~\cite{Dannecker15,GonzalezOrdiano17,Hong09}. Therefore, the input and desired output of a model  (cf. Equation~\eqref{eq. parametric_regression_with_estparam}), that for example, estimates a future time series value, $\mathsf{y}[k+H]$, using autoregressive values, i.e. $\{\mathsf{y}[k],\ldots,\mathsf{y}[k-H_{1}]\}$ -- with $H_{1} \in \mathbb{N}_{0}$ being the number of used time lags --, as well as current and past observations $\{\bm{\mathsf{u}}[k],\ldots,\bm{\mathsf{u}}[k-H_{1}]\}$ -- considered observations of various exogenous time series -- are defined in the present work as:
\begin{equation}\label{eq. time_series_forecasting_w_regression}
\begin{aligned}
y          & : = \mathsf{y}[k+H] \\
\mathbf{x} & : = \left[\mathsf{y}[k],\cdots,\mathsf{y}[k-H_{1}],\bm{\mathsf{u}}^{T}[k],\cdots,\bm{\mathsf{u}}^{T}[k-H_{1}]\right]^{T} ; k > H_{1}, k \leq K-H \text{ .}
\end{aligned}
\end{equation} 

The definition given in Equation~\eqref{eq. time_series_forecasting_w_regression} allows the creation of quantile regressions that can be used as non-parametric probabilistic forecasts.




\section{Quantile Regressions based on the Nearest Neighbors Quantile Filter}\label{sec. quant_reg_w_knn_training_set}

\begin{figure}[!hbt]
	\centering
	\captionsetup{margin=1.25cm}
	\includegraphics[width=\textwidth]{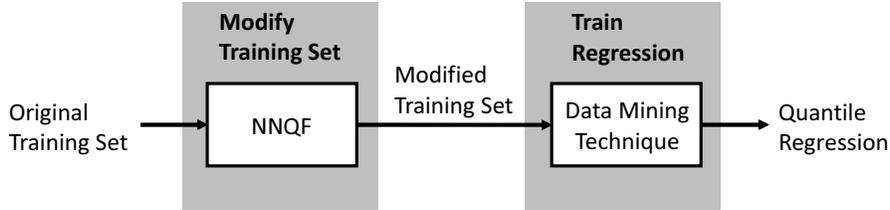}
	\caption{Creation of a quantile regression using the present contribution's approach}
	\label{fig. method_quantile_regression}
\end{figure}

As depicted in Figure~\ref{fig. method_quantile_regression}, the creation of a single quantile regression consists on modifying the available training set and using it to train a regression model with a given data mining technique and its unmodified training algorithm. 
The specifics of each step depicted in Figure~\ref{fig. method_quantile_regression} are thoroughly described in the following paragraphs. \textcolor{review1}{Additionally, the difference between the present method and the k-nearest neighbors quantile regression approach is also explained.}

\subsubsection*{Modify Training Set}

The present contribution's method starts -- just as every regression approach -- with a training set comprised of an input matrix $\mathbf{X}$ and its corresponding desired output vector $\mathbf{y}$ (cf. Equation~\eqref{eq. training_set_regression}). 
The new method assumes that the available training set is large enough to be able to accurately represent the uncertainty of $y$ given $\mathbf{x}$. This leads to the additional assumption that the output values of the nearest neighbors of $\mathbf{x}_{n}$ represent the conditional distribution of possible $y$ values given that same $\mathbf{x}_{n}$. Therefore, the new method utilizes a procedure referred to as the nearest neighbors quantile filter (NNQF) to obtain a modified training set with which a specific regression model is later to be trained. 

The NNQF begins by finding the nearest neighbors of $\mathbf{x}_{n}$, this consists in determining the nearest neighbors' index set $J^{\mathrm{pref}}_{n} \subset \{1,\dots,N\}$ that is defined as:
\begin{equation}\label{eq. nearest_neighbor_def}
\begin{aligned}
p_{\mathrm{opt}} = & \underset{p}{\operatorname{argmax}} \operatorname{card}\{ J_{n p} \}\\
\text{s.t.}~J_{n p}: & \underset{j \in J_{n p}}{\operatorname{max}} \{d(\mathbf{x}_{n},\mathbf{x}_{j})\} \leq  \operatorname{min}\{\underset{i \notin J_{n p}}{\operatorname{min}} \{d(\mathbf{x}_{n},\mathbf{x}_{i})\},\epsilon_{\mathrm{th}}\} \\
& \operatorname{card}\{J_{n p}\} \leq N_{\mathrm{NN}} \\
J^{\mathrm{pref}}_{n} = & J_{n p_{\mathrm{opt}}} \text{ ,}
\end{aligned}
\end{equation}
where $d(\cdot,\cdot)$ is the used distance function defined by a respective distance measure (as, e.g., the Euclidean distance), $\operatorname{card}\{\cdot\}$ represents the cardinality operator, $N_{\mathrm{NN}}$ denotes the number of searched nearest neighbors, and $\epsilon_{\mathrm{th}}$ is a threshold defining the maximal distance that a value $\mathbf{x}_{j}$ can have to $\mathbf{x}_{n}$ for it to be considered its nearest neighbor. In other words, finding $J^{\mathrm{pref}}_{n}$ consists in first finding various index sets $J_{n p}$ for which the following conditions hold: 
\begin{enumerate}
\item The amount of elements in $J_{n p}$ can not surpass $N_{\mathrm{NN}}$
\item The greatest distance of the nearest neighbors $\mathbf{x}_{j}; j \in J_{n p}$ to $\mathbf{x}_{n}$ cannot be greater than either the lowest distance of non-nearest neighbors $\mathbf{x}_{i}; i \notin J_{n p}$ to $\mathbf{x}_{n}$ or $\epsilon_{\mathrm{th}}$.
\end{enumerate}
Thereafter, the index $p$ of the $J_{n p}$ set containing the largest amount of elements is determined, i.e. $p_{\mathrm{opt}}$\footnote{The problem of finding more than a single $p_{\mathrm{opt}}$ can be solved by implementing the NNQF in such a way that it only finds index sets $J_{n p}$ with indexes sorted by nearest neighbors' distance in an ascending order. Furthermore, the implementation should consider only index sets $J_{n p}$ in which the indexes of nearest neighbors with the same distance are also sorted in an ascending order (cf. \ref{ap. nnqf_alg}).}. Then, the index set $J_{n p_{\mathrm{opt}}}$ is defined as the index set of the nearest neighbors, $J^{\mathrm{pref}}_{n}$.

Subsequently, the NNQF defines a vector $\mathbf{y}_{\mathrm{NN},n}$ that contains the nearest neighbors output values, i.e. $\{y_{n};n \in J^{\mathrm{pref}}_{n} \}$. Thereupon, the empirical $q$-quantile of the values inside of $\mathbf{y}_{\mathrm{NN},n}$, i.e. $\tilde{y}_{(q),n}$, is calculated using a procedure given in Definition 5 of~\cite{Hyndman96} and Method 10 in~\cite{Langford06} (cf.~\ref{ap. calc_empirical_quant}). 

After repeating the procedure for all $n = 1,\dots,N$ the NNQF defines a vector   
\begin{equation}\label{eq:new_desired_output_vector}
\tilde{\mathbf{y}}_{(q)} = [\tilde{y}_{(q),1},\cdots, \tilde{y}_{(q),N}]^{T}
\end{equation} 
that combined with the original input matrix $\mathbf{X}$ form the modified training set. \textcolor{review1}{Note that an algorithm which can be used in the implementation of the previously described NNQF is given in~\ref{ap. nnqf_alg}.}


\textbf{Remark:} An implementation of the NNQF can be divided in two main procedures: the first is the most computationally expensive, since it is comprised of the nearest neighbors calculation and of the determination of the vectors $\mathbf{y}_{\mathrm{NN},n}$, while the second part is not as computationally demanding, as it only calculates the elements of $\tilde{\mathbf{y}}_{(q)}$. Therefore if more than a single quantile regression is to be created, an optimized implementation of the NNQF is recommended. This optimized version conducts the first procedure only once. Then using the determined $\mathbf{y}_{\mathrm{NN},n}$ vectors, the optimized implementation calculates the various vectors $\tilde{\mathbf{y}}_{(q)}$ of the quantile regressions to be created. This optimized implementation is the one used in the present contribution.


\subsubsection*{Train Regression}

The previously obtained modified training set (i.e. $\mathbf{X}$ -- cf. Equation~\eqref{eq. training_set_regression} -- and $\tilde{\mathbf{y}}_{(q)}$  -- cf. Equation~\eqref{eq:new_desired_output_vector}) can then be used to train a regression model 
using a given data mining technique and its unmodified training algorithm (e.g., a linear regression trained using the least squares method). This results in a model able to estimate the conditional empirical $q$-quantile, defined by the used nearest neighbors and is given in the present contribution as
\begin{equation}\label{eq. parametric_quantile_regression_kNN}
 \hat{\tilde{y}}_{(q)} = f(\mathbf{x};\hat{\tilde{\boldsymbol{\theta}}}_{(q)}) \text{ ;}
\end{equation}
with the tilde-superscript denoting the fact that the regression is trained on the modified training set.

The described methodology has the advantages: (i) of having $q$ as a free parameter, making the creation of any quantile regression for any $q \in (0,1)$ possible, (ii) of not assuming any specific data mining technique for the creation of the quantile regression models, hence allowing the obtainment of both linear and non-linear quantile regressions without changing the original algorithms of the used data mining techniques, and (iii) of using the nearest neighbors only for the obtainment of the output vector $\tilde{\mathbf{y}}_{(q)}$, thus eliminating the necessity of saving the original training set and of conducting the nearest neighbors calculation during the application of the quantile regressions.

\subsubsection*{Difference to k-Nearest Neighbors Quantile Regression}

\textcolor{review1}{Before continuing with the description of the benchmark, it is important to clarify the main differences between the quantile regressions based on the NNQF and the classical k-nearest neighbors quantile regressions (shown for instance, in~\cite{Ma15b}). For a better comparison these differences -- in both the training and application of both types of quantile regressions -- are contained in Table~\ref{tab. diff_to_kNNRegression}.}
\begin{table}
    \begin{tabu} to \textwidth { X[0.7] | X | X }
	    & Training & Application \\ \hline
k-Nearest Neighbors Quantile Regressions & Save the available training set &  Calculate the nearest neighbors \textbf{every time} a forecast is conducted, then calculate the quantiles of the nearest neighbors' values \\ 
&& \\ \hline
Quantile Regressions based on the NNQF & Calculate the nearest neighbors \textbf{once} and train the parametric quantile regressions using the NNQF & Apply the previously trained parametric quantile regressions\\ 
\end{tabu}
\caption{\textcolor{review1}{Differences between quantile regressions based on the NNQF and the k-nearest neighbors quantile regressions}}
\label{tab. diff_to_kNNRegression}
\end{table}

\begin{figure}[!htb]
	\centering
	\includegraphics[width=\textwidth]{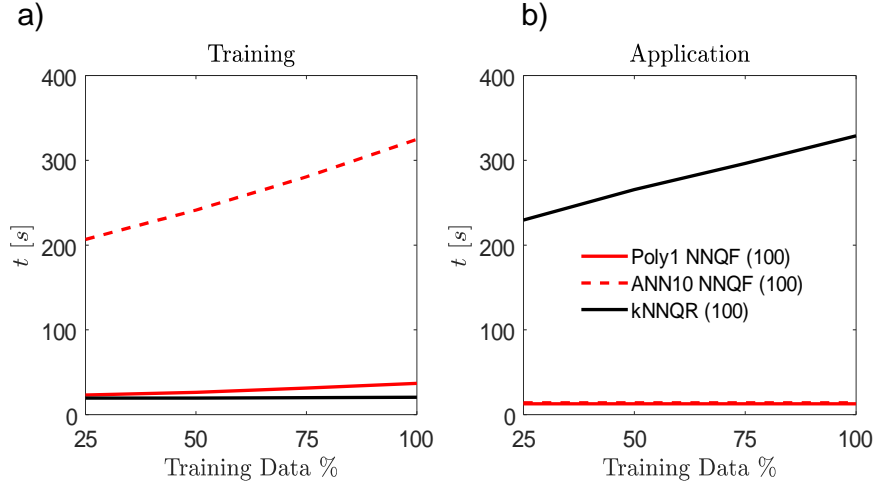}		
	\caption{\textcolor{review2}{Percentage of training data vs. time for training/applying the quantile regressions. In the legend, the number of nearest neighbors used by the nearest neighbors dependent techniques are shown in parenthesis; \textcolor{red}{Red:} with NNQF; \textbf{Black:} k-nearest neighbors quantile regression (kNNQR)}}
	\label{fig. comparison_w_kNNQR}
\end{figure}

\textcolor{review1}{As described by the contents of Table~\ref{tab. diff_to_kNNRegression}, the main advantage of using the NNQF over the more traditional k-nearest neighbors quantile regressions (kNNQRs) is the fact that the nearest neighbors are calculated only once, instead of every single time a forecast is needed. This results in the NNQF having a better scalability than the nearest neighbors quantile regressions in cases in which the training data is constantly increasing. }\textcolor{review2}{For the sake of illustration, Figure~\ref{fig. comparison_w_kNNQR} depicts the computation time needed for training and applying 99 different quantile regressions (with $q= 0.01,0.02,\dots,0.99$) using different percentages of some available training data (i.e. 25, 50, 75, and 100$\%$). To be more specific, the results stem from kNNQRs\footnote{\textcolor{review2}{In the case of the kNNQR, training and application are defined as follows: training: saving the training set; application: calculating the nearest neighbors and determining the quantile estimates}} based on~\cite{Ma15b} and NNQF-based quantile regressions obtained using a polynomial with a maximum degree of one (i.e. Poly1) or an ANN (multilayer perceptron) with 10 hidden neurons (i.e. ANN10). Additionally, all regressions use 100 nearest neighbors. Furthermore, the computer used for this example has an Intel Core i7-4790 processor with $3.6~[\mathrm{GHz}]$ and $16~[\mathrm{GB}]$ of RAM. More information about the created regressions and the data used for the example can be found in~\ref{ap. ex_regressions}.}

\textcolor{review2}{Figure~\ref{fig. comparison_w_kNNQR}a shows that the time that is necessary to train a kNNQR is more or less independent of the amount of training data used. This is an expected result, since training a kNNQR mostly consists in saving the available training set. In comparison, the regressions based on the NNQF show a different behavior, as training them requires the computation of both the nearest neighbors and the regression parameters. Henceforth, the larger the amount of available training data and the more complex the data mining technique used are, the longer the regression's training lasts. Regardless, the advantage of using the regressions based on the NNQF over kNNQRs stems from their application. As it can be observed in Figure~\ref{fig. comparison_w_kNNQR}b, the time needed for applying the quantile regressions based on the NNQF seems to be independent of the training data used. This can easily be explained by the fact that the regressions are pretrained functions that only need to be evaluated during their application. In contrast, the kNNQR requires -- as stated previously -- the computation of the nearest neighbors each time it is applied, i.e. each time a forecast is required. Hence the larger the amount of available training data is, the longer it is going to take the kNNQR to find the nearest neighbors. Even though the same can be said of the NNQF, the fact that the nearest neighbors are only searched during training and never again during application, speaks in favor of using the NNQF over the kNNQR if scalability during the application is relevant; for instance, in an online forecasting service in which the amount of training data constantly increases (e.g., the service planned for the Energy Lab 2.0~\cite{Hagenmeyer16}).}


\section{Benchmark}\label{sec. benchmark}

\subsection*{Benchmark Data}

The present contribution uses as a benchmark the data provided for the solar track of the Global Energy Forecasting Competition of 2014 (GEFCom14)~\cite{Hong16}. This dataset consists of photovoltaic (PV) power time series, $\{\mathsf{P}[k];k = 1,\dots,K \}$, -- with values normalized to values between zero and one -- of three different PV power plants and several corresponding weather forecast time series. The weather forecast time series that are relevant in the present contribution are: the surface solar radiation $\{\hat{\mathsf{G}}_{\mathrm{s}}[k];k = 1,\dots,K \}$, the surface thermal radiation $\{\hat{\mathsf{G}}_{\mathrm{th}}[k];k = 1,\dots,K \}$, and the top net solar radiation $\{\hat{\mathsf{G}}_{\mathrm{t}}[k];k = 1,\dots,K \}$\footnote{\textcolor{review1}{All radiation values used in the present article are  first order differences of the ones contained in the dataset, since all of them are originally given as accumulated values.}}. All time series are given for the time period of April $1^{st}$, 2012 to July $1^{st}$, 2014 in an hourly resolution, i.e. $K = 19704$. 

\subsection*{Benchmark Description}

\textcolor{review2}{GEFCom14 was divided into 15 tasks that consisted of a training and a test period. Also, only the last 12 tasks (i.e. Task 4 to 15) were considered for the final result~\cite{Hong16}. Table~\ref{tab. training_test_dataAmount} contains information about the data used for training and testing in each relevant task.}
\begin{table}[hbt]
\centering
\resizebox{\textwidth}{!}{
\begin{tabular}{l|ccc|ccc}
     & \multicolumn{3}{c|}{Training} & \multicolumn{3}{c}{Test} \\
Task & $K$   & $\#$ Days & Dates & $K$ & $\#$ Days & Dates  \\ \hline
4    & 10944 & 456 & 01.04.12 - 30.06.13 & 744 & 31 & 01.07.13 - 31.07.13   \\
5    & 11688 & 487 & 01.04.12 - 31.07.13 & 744 & 31 & 01.08.13 - 31.08.13  \\
6    & 12432 & 518 & 01.04.12 - 31.08.13 & 720 & 30 & 01.09.13 - 30.09.13  \\
7    & 13152 & 548 & 01.04.12 - 30.09.13 & 744 & 31 & 01.10.13 - 31.10.13  \\
8    & 13896 & 579 & 01.04.12 - 31.10.13 & 720 & 30 & 01.11.13 - 30.11.13  \\
9    & 14616 & 609 & 01.04.12 - 30.11.13 & 744 & 31 & 01.12.13 - 31.12.13  \\
10   & 15360 & 640 & 01.04.12 - 31.12.13 & 744 & 31 & 01.01.14 - 31.01.14  \\
11   & 16104 & 671 & 01.04.12 - 31.01.14 & 672 & 28 & 01.02.14 - 28.02.14  \\
12   & 16776 & 699 & 01.04.12 - 28.02.14 & 744 & 31 & 01.03.14 - 31.03.14  \\
13   & 17520 & 730 & 01.04.12 - 31.03.14 & 720 & 30 & 01.04.14 - 30.04.14  \\
14   & 18240 & 760 & 01.04.12 - 30.04.14 & 744 & 31 & 01.05.14 - 31.05.14  \\
15   & 18984 & 791 & 01.04.12 - 31.05.14 & 720 & 30 & 01.06.14 - 30.06.14  \\ \hline
\end{tabular}}
\caption{\textcolor{review1}{Amount of training and test data in each relevant task}}
\label{tab. training_test_dataAmount}
\end{table}

\textcolor{review2}{The goal at each task of the competition was to create during the training periods models able to obtain accurate quantile forecasts for the test periods~\cite{Hong16}. In the context of the present article, this is summarized as follows: first, using the data available at each training period 99 quantile regressions (i.e. $q = 0.01, 0.02, \dots, 0.99$) for each of the three PV power plants are trained using the previously presented method (cf. Section~\ref{sec. quant_reg_w_knn_training_set}) and several data mining techniques described in the next section. Note that the fact that weather forecast data was always available, but historical PV power data was only available during training periods, is taken into consideration. After their training, the regressions are used on the corresponding test period to forecast a month of each of the PV power time series on a 24 hour basis. Thereafter, the tasks' pinball-loss value averaged across all estimated quantiles is calculated, i.e.:}
\begin{equation}\label{eq. pinball_loss}
\begin{aligned}
Q_{\mathrm{PL},(q)} & = \dfrac{1}{N} \sum_{n=1}^{N}
\begin{cases}
(q-1)\cdot(y_{n}-f(\mathbf{x}_{n};\hat{\tilde{\boldsymbol{\theta}}}_{(q)})) & \text{, if } y_{n} < f(\mathbf{x}_{n};\hat{\tilde{\boldsymbol{\theta}}}_{(q)})\\
q\cdot(y_{n}-f(\mathbf{x}_{n};\hat{\tilde{\boldsymbol{\theta}}}_{(q)})) & \text{, else}
\end{cases} \\
Q_{\mathrm{PL}} & = \operatorname{mean}\{Q_{\mathrm{PL},(0.01)},Q_{\mathrm{PL},(0.02)},\cdots, Q_{\mathrm{PL},(0.99)}\} \text{ .}
\end{aligned} 
\end{equation}
\textcolor{review1}{The closer $Q_{\mathrm{PL}}$ gets to zero, the more accurate the quantile regressions are}. 

After evaluating all 12 relevant tasks the average pinball-loss across all of them is determined and compared to the average obtained by \textcolor{review1}{the winner of GEFCom14 (i.e. $1.21\%$)}, which used a combination of gradient boosting and nearest neighbors regressions to solve the different tasks~\cite{Huang16}. 

\textcolor{review1}{Additionally, the necessary time $t$ for the training, application, and both of all obtained quantile regressions in Task 15 is measured and multiplied to the number of processing cores $N_{\mathrm{cores}}$  and to the used processors' clock rate $f_{\mathrm{clock}}$. This results in dimensionless values representing the amount of clock cycles -- in the present work referred to as the computational effort -- under the assumption that all cores are being used at their maximum potential:}
\begin{equation}\label{eq. processor_time}
\begin{aligned}
C  & = t\cdot N_{\mathrm{cores}}\cdot f_{\mathrm{clock}} \text{ ,} \\
\end{aligned}
\end{equation}
\textcolor{review1}{with $C$ representing the computational effort. Additionally, the computer used in the present contribution has $N_{\mathrm{cores}} = 8$ and $f_{\mathrm{clock}} = 3.6~[\mathrm{GHz}]$. The obtained values are compared to the values calculated using information reported by the winner of GEFCom14, i.e. $t = 662~[\mathrm{s}]$ for training, $t = 96~[\mathrm{s}]$ for the application, $N_{\mathrm{cores}} = 256$ and $f_{\mathrm{clock}} = 2.6~[\mathrm{GHz}]$~\cite{Huang16}}\footnote{The procedures given by the winner of GEFCom14 in~\cite{Huang16} that were considered to be the training and application of their models are: Training: preprocessing, modelling, and boosting. Applying: PDF estimation step}. \textcolor{review1}{A comparison solely based on the computation time is considered to be unfair, due to the difference in power between the present article's computer and that of the GEFCom14 winning team. Therefore, it is assumed that using a metric like the computational effort, which takes into consideration the number of processing cores and the clock rates, results in a more accurate comparison.} \textcolor{review2}{Note that two other data mining techniques, which are described in the next section, are also used as benchmark for comparison.}

\textcolor{review2}{Furthermore, to assess how well the NNQF-based regressions would have performed during the actual competition, a skill score, $Q_{\mathrm{Sk}}$, is also used during the evaluation. The value of $Q_{\mathrm{Sk}}$ describes the improvement, in terms of pinball-loss, achieved in relation to the competition's benchmark, i.e.:}
\begin{equation}\label{eq. skillScore}
\begin{aligned}
Q_{\mathrm{Sk}} = & \dfrac{Q_{\mathrm{PL,B}}-Q_{\mathrm{PL}}}{Q_{\mathrm{PL,B}}} \text{ ;}
\end{aligned}
\end{equation} 
\textcolor{review2}{with $Q_{\mathrm{PL,B}}$ representing the pinball-loss of GEFCom14's benchmark.}

\textcolor{review2}{Another value that has to be taken into consideration when evaluating quantile regressions is their reliability~\cite{Gneiting07,Pinson07}. The reliability describes if the percentage of values that are equal to or lower than the outputs of a given quantile regression are actually close to the desired probability $q$. The percentage of values necessary to asses a quantile regression's reliability is calculated in the present article as:} 
\begin{equation}\label{eq. reliabilityDeviation}
\begin{aligned}
Q_{\mathrm{R},(q)} = & \dfrac{1}{N}\underset{n=1,\dots,N}{\operatorname{card}}\{y_{n} \leq f(\mathbf{x}_{n};\hat{\tilde{\boldsymbol{\theta}}}_{(q)})\}\text{ ;}
\end{aligned}
\end{equation}

\textcolor{review2}{with $Q_{\mathrm{R},(q)}$ being the percentage of values equal to or lower than the outputs of a given quantile regression and $\operatorname{card}\{\cdot\}$ representing the cardinality operator. It is important to consider that the reliability evaluation on PV power has to be conducted only on day values, since taking into account the trivial values at night skews the result. For such reason, the $Q_{\mathrm{R},(q)}$ is calculated only on a subset of values for which the normalized power or the estimated median (i.e. $\hat{\tilde{y}}_{(0.5)}$) is larger than 0.05. Doing so eliminates the risk of taking night values into account. Moreover, the normalized power value of 0.05 is chosen as threshold to avoid considering cloudy day values as night measurements.}

\subsection*{Data Mining Techniques}

The quantile regressions are created using the presented method (cf. Section~\ref{sec. quant_reg_w_knn_training_set}) and different data mining techniques: The first are polynomials (i.e. multiple linear regressions) with a maximally allowed degree of one up to four, which are referred to as Poly1-4. The final two techniques are ANNs (multilayer perceptrons), both with a single hidden layer but a different number of hidden neurons; the first, ANN6, has six and the second, ANN10, has ten. In addition, different $N_{\mathrm{NN}}$ values -- i.e. 50, 100, 150, and 200 -- are used for the calculation of the desired output vector $\tilde{\mathbf{y}}_{(q)}$ (cf. Equation~\eqref{eq:new_desired_output_vector}) with which the models of the different techniques are to be trained\footnote{The value of $\epsilon_{\mathrm{th}}$ (cf. Equation~\eqref{eq. nearest_neighbor_def}) was set to $\mathrm{Inf}$ in the implementation of NNQF, to allow the NNQF to always find for each input vector $N_{\mathrm{NN}}$ nearest neighbors.}. 
Furthermore, all quantile regressions are created using the open-source MATLAB toolbox SciXMiner\footnote{\textcolor{review2}{Notice that the GEFCom14 winner's method used for comparison is implemented in R~\cite{Huang16}.}}~\cite{Mikut17}. \textcolor{review1}{Note that the techniques being used are quite simple, since the goal of the current article is not to present a novel forecasting model, but rather to show the possibility of training quantile regressions based on the NNQF (i.e. without directly minimizing the sum of pinball-losses).}

All of the present contribution's quantile regressions are created using only values of forecast weather time series as input; i.e. values from the dataset's surface solar radiation, the surface thermal radiation, and the top net solar radiation time series. The lack of autoregressive values as input comes from the fact that the PV power time series were never updated during the test periods of the competition. To assure the consistency between the currently presented benchmark and Section~\ref{sec. quant_reg_w_knn_training_set}, Equation~\eqref{eq. time_series_forecasting_w_regression} is reformulated as 
\begin{equation}\label{eq. pv_power_time_series_forecasting_w_regression}
\begin{aligned}
y             & : = \mathsf{P}[k+H] ; H = 24\\
\mathbf{x}    & : = \left[\bm{\mathsf{u}}^{T}[k],\cdots,\bm{\mathsf{u}}^{T}[k-H_{1}]\right]^{T} ; k > H_{1}; k \leq K-H; H_{1} = 24 \\
\bm{\mathsf{u}}[k] & : = \left[\hat{\mathsf{G}}_{\mathrm{s}}[k+H],\hat{\mathsf{G}}_{\mathrm{th}}[k+H],\hat{\mathsf{G}}_{\mathrm{t}}[k+H]\right]^{T} \text{ .}
\end{aligned}
\end{equation}
The use of $\hat{\mathsf{G}}_{\mathrm{s}}[k+H]$, $\hat{\mathsf{G}}_{\mathrm{th}}[k+H]$, and $\hat{\mathsf{G}}_{\mathrm{t}}[k+H]$ stems from the fact that the values were always available at the moment of conducting the forecast during the competition.

To reduce the dimensionality of the input space and the probability of over-fitting the models, the four most relevant features inside of $\mathbf{x}$ are selected -- via a forward feature selection -- and normalized (to values between 0 and 1), prior to the nearest neighbors approach and to the obtainment of the quantile regressions. For the sake of simplicity, the input vector containing the four selected and normalized features is further referred to as $\mathbf{x}$. \textcolor{review1}{Moreover, the distance function used during the calculation of the nearest neighbors is the weighted Euclidean distance with the inverse of the selected normalized features' variance as weights}.

Also, to reduce the possibility of quantile crossing and to assure positive power forecasts, the polynomial quantile regressions are trained using the following constrained least squares method:
\begin{equation}
\begin{aligned}
& \underset{\tilde{\boldsymbol{\theta}}_{(q)}}{\operatorname{argmin}} &  \sum_{n=1}^{N} (\tilde{y}_{(q),n} & - f(\mathbf{x}_{n};\tilde{\boldsymbol{\theta}}_{(q)}))^{2}\\
& \operatorname{s.t.} & & \begin{cases}
 f(\mathbf{x};\tilde{\boldsymbol{\theta}}_{(q)})  \geq 0  & \text{ if $q$ = 0.01}\\
 f(\mathbf{x};\tilde{\boldsymbol{\theta}}_{(q)})  \geq f(\mathbf{x};\tilde{\boldsymbol{\theta}}_{(q-0.01)}) & \text{ else}  \text{ .}
\end{cases} 
\end{aligned}
\end{equation}
Moreover, to further avoid quantile crossing all quantile regressions -- including the ones created by the ANNs -- are subject to similar constraints during their application, i.e.
\begin{equation}
\hat{\tilde{y}}_{(q)} = \begin{cases}
				\operatorname{max}(f(\mathbf{x};\hat{\tilde{\boldsymbol{\theta}}}_{(q)}), 0) & \text{ if $q = 0.01$} \\
				\operatorname{max}(f(\mathbf{x};\hat{\tilde{\boldsymbol{\theta}}}_{(q)}), f(\mathbf{x};\hat{\tilde{\boldsymbol{\theta}}}_{(q-0.01)})) & \text{ else} \text{ .}			
				\end{cases} 
\end{equation}

\textcolor{review2}{In addition to the GEFCom14 winner, other techniques are also used as benchmark. The first are k-nearest neighbors quantile regressions (i.e. kNNQR) that use 50, 100, 150, or 200 nearest neighbors and are based on the method described in~\cite{Ma15b}. The second are traditional quantile regressions based on polynomial models with maximum allowed degrees of one up to four (i.e. Poly1-4 TQR). In the present context, traditional refers to the fact that the regressions are trained by minimizing the sum of pinball-losses (cf. Equation~\eqref{eq. pinball_loss_opt})\footnote{\textcolor{review2}{The training algorithm of these regressions minimizes the sum of pinball-losses~\cite{Koenker05} using a MATLAB implementation found in \href{https://de.mathworks.com/matlabcentral/fileexchange/32115-quantreg-x-y-tau-order-nboot-}{de.mathworks.com/matlabcentral/fileexchange/32115-quantreg-x-y-tau-order-nboot-}, which has been modified to avoid quantile crossing.}}. Lastly, both the input vector and desired output of these other types of regressions are the ones defined in Equation~\eqref{eq. pv_power_time_series_forecasting_w_regression}.}

As a final remark, it needs to be mentioned that PV power values with $\hat{G}_{\mathrm{s}}[k+H] \leq 100000~[\mathrm{Jh^{-1}m^{-2}}]$ are considered to be night values and thus are eliminated from the utilized training sets and automatically set to zero during the test periods.

\section{Results and Discussion}\label{sec. results}

\textcolor{review2}{Table~\ref{tab. relative_pinball_loss} contains the average pinball-loss (averaged over all relevant tasks) of all tested data mining techniques and all used benchmarks, including the winner of GEFCom14.}
\begin{table}[hbt]
\centering
\footnotesize
\begin{tabular}{|l|cccc|}
\hline
$N_{\mathrm{NN}}$ & 50 & 100 & 150 & 200  \\ \hline
& \multicolumn{4}{c|}{ $Q_{\mathrm{PL}}~[\%]$ }   \\ \hline
\textbf{with NNQF} & & & & \\ \hline
Poly1 & 1.93 & 1.92 & 1.90 & 1.90   \\
Poly2 & 1.94 & 1.93 & 1.93 & 1.92  \\
Poly3 & 1.92 & 1.92 & 1.91 & 1.91  \\
Poly4 & 1.94 & 1.93 & 1.93 & 1.92  \\
ANN6  & 1.55 & 1.57 & 1.58 & 1.61  \\
ANN10 & 1.51 & 1.56 & 1.57 & 1.59  \\ \hline
\textbf{Benchmarks} & & & & \\  \hline
kNNQR & 1.81 & 1.89 & 1.94 & 1.99 \\
Poly1 TQR & \multicolumn{4}{c|}{1.81}  \\
Poly2 TQR & \multicolumn{4}{c|}{1.76}  \\
Poly3 TQR & \multicolumn{4}{c|}{1.76}  \\
Poly4 TQR & \multicolumn{4}{c|}{1.76}  \\ 
GEFCom14 Winner & \multicolumn{4}{c|}{1.21} \\ \hline
\end{tabular}
\caption{\textcolor{review2}{Average pinball-loss; the techniques showing only one result are the ones that do not use nearest neighbors}}
\label{tab. relative_pinball_loss}
\end{table}
%

\begin{figure}[!htb]
	\centering
	\includegraphics[width=\textwidth]{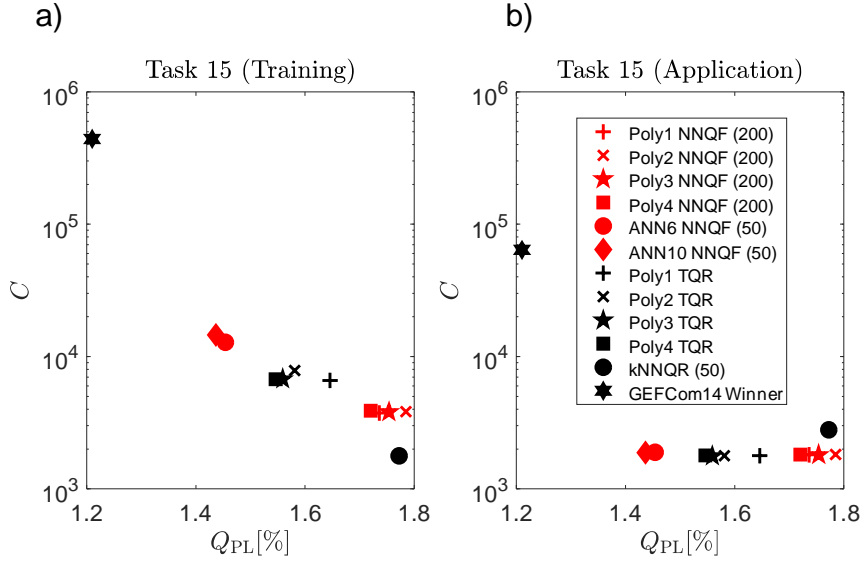}	
	\caption{Pinball-loss vs. computational effort for training/applying the quantile regressions. In the legend, the number of nearest neighbors used by the nearest neighbors dependent techniques are shown in parenthesis; \textcolor{red}{Red:} with NNQF; \textbf{Black:} benchmarks}
	\label{fig. time_vs_pinball}
\end{figure}

The obtained low pinball-loss values -- ranging from $1.51\%$ to $1.94\%$ -- confirm that the presented method allows the creation of quantile regressions independently of the utilized data mining technique. Furthermore, the differences between the pinball-loss of the GEFCom14 winner (i.e. $1.21\%$) and the NNQF-based regressions are all below $1\%$ and range from $0.73\%$ when using Poly2 ($N_{\mathrm{NN}} = 50$) to $0.30\%$ in the case of ANN10 ($N_{\mathrm{NN}} = 50$). \textcolor{review1}{Moreover, it seems that an increasing number of nearest neighbors improves the quantile regressions trained using polynomial models, but decreases the accuracy of the ones trained using artificial neural networks. \textcolor{review2}{Also, the results show that the polynomial regression based on the NNQF perform generally worse than their TQR counterparts. This does not come as a surprise, since the latter are actually trained to minimize the sum of pinball-losses. 
Regardless, both ANNs perform better than every polynomial and kNNQR model, demonstrating the fact that ANN quantile regressions can be trained easily with their traditional gradient based algorithms, since the non-differentiable sum of pinball-losses is not used. This is one of the advantages of using the present contribution's method.} Nevertheless, none of the NNQF-based regressions are able to perform better than the approach of the winner of GEFCom14, which may be attributed to the extremely simple data mining techniques being used in the present article. Still, the advantage of using the NNQF becomes clear when considering the computational effort. \textcolor{review2}{For the sake of illustration, Figure~\ref{fig. time_vs_pinball} depicts two separate plots of the average pinball-loss values of Task 15 and their respective computational effort for either training or applying the models.} \textcolor{review2}{Note that the values depicted for the nearest neighbors dependent techniques are the ones obtained with the number that resulted in their best average pinball-loss (i.e. $N_{\mathrm{NN}} =200$ for the polynomials and $N_{\mathrm{NN}} =50$ for the ANNs and kNNQR). For more information of the computation times in seconds, readers are referred to~\ref{ap. tr_ap_times} Table~\ref{tab. explicit_times}.}}


\textcolor{review2}{At a first glance, Figure~\ref{fig. time_vs_pinball} shows that while the differences in pinball-loss between the present paper's techniques and the winner's method remain small, the differences in computational effort do vary considerably, since all regressions based on the nearest neighbors quantile filter (i.e. the NNQF) require a much lower computational effort. For instance, ANN10 with $N_{\mathrm{NN}} = 50$ requires around 30 times less effort for both training and application.} 

\textcolor{review2}{In turn, Figure~\ref{fig. time_vs_pinball}a shows that the TQRs have better $Q_{\mathrm{PL}}$ values than the NNQF-based polynomial models, but require a more computationally demanding training. This can be explained by the fact that the TQRs are trained using the sum of pinball-losses, which is more difficult to minimize than the sum of squared errors used by the NNQF-based polynomials. Henceforth, the use of the NNQF can be seen as a trade-off between the training time required (reduced by avoiding the minimization of the sum of pinball-losses) and the accuracy in terms of pinball-loss. Figure~\ref{fig. time_vs_pinball}a also demonstrates that the NNQF allows the training of acceptable quantile regressions based on more complex techniques (e.g., the ANNs) without drastically increasing the computational effort. This is of importance, as such techniques may provide better forecasts than simpler approaches, just as in the current example. Also, the previously mentioned trade-off and the possibility of easily training quantile regressions with more complex techniques may be extremely relevant to particular situations, for instance, when implementing the regressions as part of an online forecasting service requiring the fast computation of probabilistic forecasts with not so powerful computers. 
In addition, the fact that the NNQF is the same regardless of the utilized data mining technique shows that the difference in the $C$ values depicted in Figure~\ref{fig. time_vs_pinball}a stems from the different training algorithms and not from the filtering step. 
Furthermore, the reason behind the kNNQR showing the lowest training effort is that its training consists only in saving the available training data.} 

\textcolor{review2}{Figure~\ref{fig. time_vs_pinball}b shows that all techniques with the exception of the kNNQR and the GEFCom14 winner require a similar computational effort for their application. The reason thereof is that the application of all NNQF-based regressions and TQRs consists in evaluating the previously trained functions. In contrast, the kNNQR and GEFCom14 winner's method have to compute the nearest neighbors every time a new forecast is conducted, as they are a type of nearest neighbors regressions. This not only increases the corresponding $C$ value, but also causes the effort to increase proportionally with an increasing amount of training data (as shown in Section~\ref{sec. quant_reg_w_knn_training_set} Figure~\ref{fig. comparison_w_kNNQR}); a property that is detrimental for the implementation of this type of regressions in an online forecasting service in which the available data constantly increases.} 

\begin{figure}[!htb]
	\centering
	\captionsetup{margin=1.25cm}
	\includegraphics[width=\textwidth]{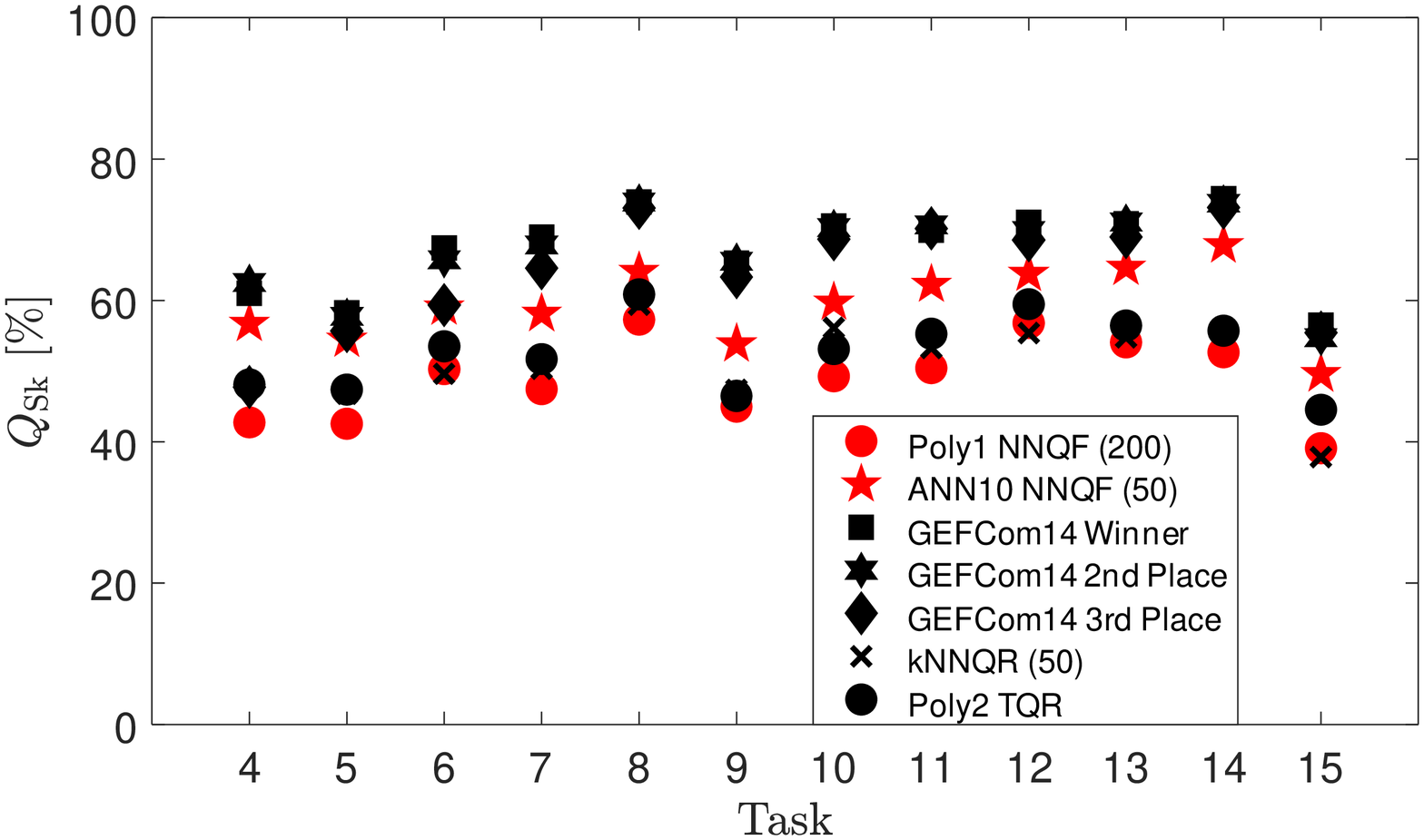}		
	\caption{\textcolor{review2}{Pinball-loss improvement relative to the GEFCom14 benchmark. In the legend, the number of nearest neighbors used by the nearest neighbors dependent techniques are shown in parenthesis; \textcolor{red}{Red:} with NNQF; \textbf{Black:} benchmarks}}
	\label{fig. comparison_3best_wBenchmark_relative}
\end{figure}

\textcolor{review2}{Another interesting comparison can be seen in Figure~\ref{fig. comparison_3best_wBenchmark_relative}, which shows the skill score defined in Equation~\eqref{eq. skillScore} that is achieved by the best performing NNQF-based polynomial and ANN regressions (cf. Table~\ref{tab. relative_pinball_loss}). Additionally, the skill scores of the first~\cite{Huang16}, second~\cite{Nagy16}, and third~\cite{Juban16} places of the competition and the best performing TQR and kNNQR (cf. Table~\ref{tab. relative_pinball_loss}) are also depicted. The goal of Figure~\ref{fig. comparison_3best_wBenchmark_relative} is to show the performance that the NNQF-based regressions would have had during the actual competition in relation to the GEFCom14 benchmark. Readers interested in the pinball-loss values obtained on each relevant task by all plotted methods, as well as the GEFCom14 benchmark are referred to~\ref{ap. tr_ap_times} Table~\ref{tab. explicit_pinball_loss}.}

\textcolor{review2}{In general, Figure~\ref{fig. comparison_3best_wBenchmark_relative} shows that the NNQF-based quantile regressions are able to obtain better results than the benchmark used during GEFCom14. Furthermore, while Poly1 delivers -- as expected due to the trade-off mentioned previously -- slightly worse and sometimes similar results than the TQR and kNNQR benchmarks, the NNQF allows ANN10 to perform better than them on all relevant tasks. Additionally, the use of the ANN10 NNQF-based regressions in the first three relevant tasks resulted in improvements that are similar to those of the competition's winners. After Task 6, nonetheless, the ANN10 regressions obtained slightly worse results than the competition's best three places, which may be caused (i) by the fact that the ANN10 may be a simple technique in comparison, (ii) by the previously discussed trade-off that results from using the NNQF, and (iii) by the fact that the competition participants improved their approaches after each task (cf.~\cite{Huang16}), while the structure of the present articles models remained the same. Unfortunately, a comparison based on the computational effort of the second and third place is not possible, since the necessary information is not provided by the authors. Likewise, an accurate re-implementation of the first three places' methods is considered unfeasible, as the implementation is not thoroughly described nor the code is provided.}

\begin{figure}[!htb]
	\centering
	\captionsetup{margin=1.25cm}
	\includegraphics[width=\textwidth]{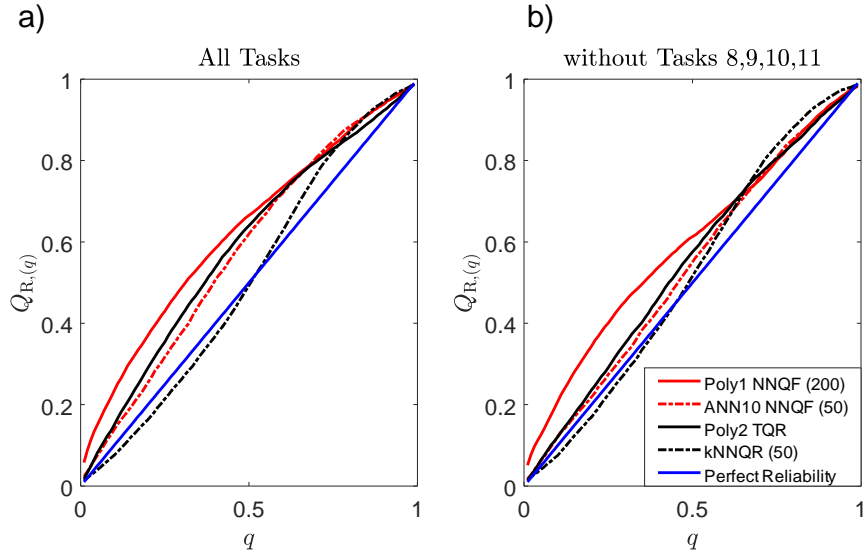}		
	\caption{\textcolor{review2}{Reliability results obtained on the GEFCom14 data. In the legend, the number of nearest neighbors used by the nearest neighbors dependent techniques are shown in parenthesis; \textcolor{red}{Red:} with NNQF; \textbf{Black:} benchmarks; \textcolor{blue}{Blue:} perfect reliability}}
	\label{fig. reliabilityDev}
\end{figure}

\textcolor{review2}{Finally, Figure~\ref{fig. reliabilityDev} depicts two reliability plots. The plots show the $Q_{\mathrm{R},(q)}$ values (cf. Equation~\eqref{eq. reliabilityDeviation}) plotted against the desired probability $q$ of the best performing NNQF-based polynomial and ANN quantile regressions in Table~\ref{tab. relative_pinball_loss}. Additionally, the reliability plots of the kNNQR and TQR benchmarks that perform best according to Table~\ref{tab. relative_pinball_loss} are also shown. Furthermore, the lack of information provided in~\cite{Huang16} regarding the reliability of the GEFCom14 winner's method is the reason for the absence of the corresponding curve in the plot.}

\textcolor{review2}{Figure~\ref{fig. reliabilityDev}a shows first the reliability plots over all relevant tasks of the competition. At a first glance, the kNNQR appears to be the most reliable technique, while all other ones seem to overestimate the desired values. However, an investigation of the results shows, that the NNQF-based regressions obtain abnormally large $Q_{\mathrm{R},(q)}$ values in a period that consists of Tasks 8, 9, 10, and 11 (i.e. the period that can be considered summer in the region where the PV systems are located; cf. Table~\ref{tab. training_test_dataAmount}). After removing the period in question and calculating again the percentages, the results of both the NNQF-based regressions and the TQR benchmark improve. Figure~\ref{fig. reliabilityDev}b shows that while Poly1 remains the technique with the least reliable regressions (especially considering the lowest quantiles), ANN10 obtains $Q_{\mathrm{R},(q)}$ values that are not only close to the desired probability, but also similar and in some cases better than those of the TQR and kNNQR benchmark. Considering the fact that all NNQF-based regressions as well as all TQR models are trained to map forecast radiation values to quantiles of the future PV power directly (cf. Equation~\eqref{eq. pv_power_time_series_forecasting_w_regression}), it can be argued that an overestimation of the solar radiation during the period of Tasks 8 to 11 is a possible cause of the large $Q_{\mathrm{R},(q)}$ values obtained. This is further supported by the fact, that the kNNQR results remain more or less the same after removing the period in question, as the method does not estimate the quantiles with the weather forecasts directly, but rather uses them to find similar past days from which their generated power is then used to estimate the desired quantiles. In other words, past generated power is mapped to future power quantiles, meaning that an overestimation of the solar radiation does not necessarily affect its results. The possible preprocessing of the forecast solar radiation values to reduce the problematic described previously is out of the scope of the present article, yet is to be studied in future related works.}

\section{Conclusion \& Outlook}\label{sec. conclusion}

\textcolor{review1}{The present work describes the new nearest neighbors quantile filter (NNQF), i.e. a procedure for the obtainment of a modified training set that allows the creation of parametric quantile regressions without the minimization of the non-differentiable sum of pinball-losses. The use of the NNQF not only eliminates the risks of training quantile regressions with gradient based optimization, but also opens the possibility of training quantile regressions using ``out of the box'' training algorithms (i.e. training algorithms found in traditional statistical and machine learning libraries) without modifying them. The use of the NNQF is validated on the solar track dataset of the Global Energy Forecasting Competition of 2014 (GEFCom14). The obtained results show that the present contribution's new method succeeds at creating quantile regressions with a similar accuracy as the winners of the GEFCom14 competition; as the differences between the obtained pinball-loss values and those of the winners are all below $1\%$. Furthermore, the computational effort (i.e. a value defined to take into consideration the computational time and the power of the used computer) necessary for solving a task of the competition is shown to be 30 times lower than that of the winner of GEFCom14. \textcolor{review2}{Additionally, the comparison of the results stemming from the NNQF-based regressions and those of the winner and two other benchmarks, provide evidence of the trade-off that seems to exist between computational effort and pinball-loss accuracy. In general, the results support the possibility of using the presented method to obtain quantile regressions (i.e. probabilistic forecasts) quickly and without the need of an extremely powerful computer.} This outcome may be beneficial when implementing quantile regressions based on the NNQF as part of an online forecasting service in which extremely powerful machines may not be necessarily available. Regardless of the obtained results, there are still a number of issues that need to be investigated further. For instance, the effects that both the threshold parameter (i.e. $\epsilon_{\mathrm{th}}$) and/or the amount of training data may have on the results, the possibility of obtaining probabilistic forecasts for load and wind power time series using the presented approach, and the creation of quantile regressions based on the NNQF using more sophisticated data mining techniques. \textcolor{review2}{Moreover, approaches able to mitigate the effects that over-/underestimating weather forecast may have on the regressions reliability have to be investigated further.} Additionally, the implementation of quantile regressions based on the NNQF in the forecasting service of the Energy Lab 2.0~\cite{Duepmeier15,Hagenmeyer16} needs also to be studied in future related works.}


\section*{Acknowledgments}
The present contribution is supported by the Helmholtz Association under the Joint Initiative ``Energy System 2050 - A Contribution of the Research Field Energy".

\appendix
\section{Calculation of the Empirical Quantile} \label{ap. calc_empirical_quant}
The calculation of the empirical quantile $\tilde{y}_{(q),n}$ of the values contained inside of $\mathbf{y}_{\mathrm{NN},n}$ using Definition 5 of~\cite{Hyndman96} and Method 10 in~\cite{Langford06} (for the case of finding $N_{\mathrm{NN}}$ nearest neighbors) begins by defining every value 
\begin{equation}
\mathbf{y}_{\mathrm{NN},n} = [y_{\mathrm{NN},n1},\cdots, y_{\mathrm{NN},ni} ,\cdots, y_{\mathrm{NN},nN_{\mathrm{NN}}}]^{T} \text{ ,}
\end{equation}
as quantiles with corresponding probabilities
\begin{equation}
[q_{\mathrm{NN},n1},\cdots,q_{\mathrm{NN},nN_{\mathrm{NN}}}]^{T} =\left[\dfrac{0.5}{N_{\mathrm{NN}}},\dfrac{1.5}{N_{\mathrm{NN}}},\cdots,\dfrac{N_{\mathrm{NN}}-0.5}{N_{\mathrm{NN}}}\right]^{T} \text{ .}
\end{equation}
Afterwards, the desired empirical quantile $\tilde{y}_{(q),n}$ can be calculated via linear interpolation.
\begin{equation}
\begin{aligned}
\tilde{y}_{(q),n} = 
& \sum_{i = 2}^{N_{\mathrm{NN}}} \Bigg(\left[\dfrac{y_{\mathrm{NN},ni}-y_{\mathrm{NN},n(i-1)}}{q_{\mathrm{NN},ni}-q_{\mathrm{NN},n(i-1)}} (q-q_{\mathrm{NN},n(i-1)}) + y_{\mathrm{NN},n(i-1)} \right] \\
& \cdot I(q_{\mathrm{NN},n(i-1)} \leq q < q_{\mathrm{NN},ni})\Bigg) + y_{\mathrm{NN},n1}\cdot I(q < q_{\mathrm{NN},n1})\\
& + y_{\mathrm{NN},nN_{\mathrm{NN}}}\cdot I(q_{\mathrm{NN},nN_{\mathrm{NN}}}\leq q) \text{ , } \\
\end{aligned} 
\end{equation}
with $I(\cdot)$ representing an indicator function that is one if its condition is fulfilled and zero otherwise. 

\section{Nearest Neighbors Quantile Filter Algorithm}\label{ap. nnqf_alg}
An algorithm that can be used to apply the nearest neighbors quantile filter (cf. Equation~\eqref{eq. nearest_neighbor_def}) is presented in Algorithm~\ref{alg. kNN_quant_reg_desired_output}.

\begin{algorithm}[!hbt]
\caption{Nearest neighbors quantile filter}\label{alg. kNN_quant_reg_desired_output}
\begin{algorithmic}[1]
\Function{NNQF}{$\mathbf{y}$,$\mathbf{X}$,$N_{\mathrm{NN}}$,$q$,$\epsilon_{\mathrm{th}}$}
\State $N \gets$ number of training samples defined by the length of $\mathbf{y}$
\State Preallocate $N \times 1$ vector $\tilde{\mathbf{y}}_{(q)}$ with all its elements $\tilde{y}_{(q),n}$ set to zero
\For{ $n = 1$ to $N$ }
\For{ $j = n+1$ to $N$ }
\State $d_{nj} \gets $ distance of $x_{n}$ to $x_{j}$
\EndFor
\State $J^{\mathrm{pref}}_{n} \gets$ indexes of $x_{j}$ vectors with $d_{nj} \leq \epsilon_{\mathrm{th}}$ sorted by increasing distance
\State Sort the elements of $J^{\mathrm{pref}}_{n}$ with same $d_{nj}$ values in ascending order
\If{$\operatorname{card}\{J^{\mathrm{pref}}_{n}\} > N_{\mathrm{NN}}$}
\State $~J^{\mathrm{pref}}_{n} \gets$ the first $N_{\mathrm{NN}}$ elements of $J^{\mathrm{pref}}_{n}$
\EndIf
\State $\tilde{y}_{(q),n} \gets$ empirical $q$-quantile of the values $\{y_{n};n \in J^{pref}_{n} \}$
\EndFor
\State \textbf{return} $\tilde{\mathbf{y}}_{(q)}$
\EndFunction
\end{algorithmic}
\end{algorithm}

\section{Data used in Figure~\ref{fig. comparison_w_kNNQR}}\label{ap. ex_regressions}

\textcolor{review2}{The data used to obtained the results in Figure~\ref{fig. comparison_w_kNNQR} stems from a simulated load time series (referred to in the current Appendix as $\{P[k];k=1,\dots,K\}$) obtained using the same load benchmark generator as in~\cite{Waczowicz18_Diss,Klaiber17}. The time series contains three years of hourly load measurements (i.e. $K = 26280$) of a simulated household. Additionally, half of the time series is used as training set and half of it is used as test set. Furthermore, the desired output and input vector used for the regressions in the example are defined as follows:}
\begin{equation}
\begin{aligned}
y & = \mathsf{P}[k+H]; H = 24 \\
\mathbf{x} & = [P[k],\cdots,P[k-H_{1}]]^{T}; k > H_{1}; k\leq K-H; H_{1} = 168 \text{.} \\
\end{aligned}
\end{equation}
\textcolor{review2}{In other words, the regressions are trained to forecast load values 24 hours in the future using the information of the past week as input.}

\section{Results Additional Information}\label{ap. tr_ap_times}

Table~\ref{tab. explicit_times} contains the time in seconds of the NNQF-based quantile regressions and benchmarks that are trained and applied with the computer used for the present contribution, which has an Intel Core i7-4790 processor with $3.6~[\mathrm{GHz}]$ and $16~[\mathrm{GB}]$ of RAM.
\setcounter{table}{0}
\begin{table}[!hbt]
\centering
\begin{tabular}{|l|cccc|cccc|}
\hline
$N_{\mathrm{NN}}$ & 50 & 100 & 150 & 200 & 50 & 100 & 150 & 200 \\ \hline
& \multicolumn{4}{c|}{ $t~[\mathrm{s}]$ (Training) } & \multicolumn{4}{c|}{ $t~[\mathrm{s}]$ (Application) } \\ \hline
\textbf{with NNQF} & \multicolumn{8}{c|}{} \\ \hline
Poly1 & 121 & 124 & 126 & 129 & 62 & 63 & 63 & 63 \\
Poly2 & 124 & 128 & 131 & 133 & 63 & 63 & 63 & 63 \\
Poly3 & 124 & 126 & 130 & 132 & 63 & 63 & 63 & 63 \\
Poly4 & 125 & 129 & 132 & 135 & 63 & 62 & 63 & 63 \\
ANN6  & 445 & 449 & 461 & 453 & 66 & 65 & 65 & 65 \\
ANN10 & 505 & 505 & 511 & 518 & 65 & 66 & 69 & 65 \\ \hline
\textbf{Benchmarks} & \multicolumn{8}{c|}{} \\ \hline
kNNQR & 62 & 62 & 62 & 62 & 97 & 97 & 97 & 97 \\
Poly1 TQR & \multicolumn{4}{c|}{229} & \multicolumn{4}{c|}{62}  \\
Poly2 TQR & \multicolumn{4}{c|}{272} & \multicolumn{4}{c|}{62}  \\
Poly3 TQR & \multicolumn{4}{c|}{235} & \multicolumn{4}{c|}{62}  \\
Poly4 TQR & \multicolumn{4}{c|}{234} & \multicolumn{4}{c|}{62}  \\ \hline
\end{tabular}
\caption{Time (in seconds) for training and applying quantile regressions in Task 15}
\label{tab. explicit_times}
\end{table}


\textcolor{review2}{Table~\ref{tab. explicit_pinball_loss} contains the pinball-loss that is obtained on each relevant task by the NNQF-based regressions with the best average results (cf. Table~\ref{tab. relative_pinball_loss}), the competition's first, second, and third place, and the GEFCom14 benchmark.}
\begin{table}[!hbt]
\centering
\resizebox{\textwidth}{!}{
\begin{tabular}{|l|cccccccccccc|}
\hline
Tasks & 4 & 5 & 6 & 7 & 8 & 9 & 10 & 11 & 12 & 13 & 14 & 15\\ \hline
 & \multicolumn{12}{c|}{$Q_{\mathrm{PL}}~[\%]$}\\ \hline
\textbf{with NNQF} & \multicolumn{12}{c|}{} \\ \hline
Poly1 $N_{\mathrm{NN}} = 200$ & 1.90 & 2.23 & 1.78 & 1.90 & 2.05 & 1.96 & 2.14 & 1.98 & 1.88 & 1.73 & 1.52 & 1.74 \\
ANN10 $N_{\mathrm{NN}} = 50$  & 1.43 & 1.77 & 1.47 & 1.51 & 1.72 & 1.65 & 1.70 & 1.51 & 1.58 & 1.33 & 1.03 & 1.44 \\ \hline
\textbf{Benchmarks} & \multicolumn{12}{c|}{} \\ \hline
kNNQR $N_{\mathrm{NN}} = 50$ & 1.73 & 2.07 & 1.81 & 1.79 & 1.94 & 1.88 & 1.85 & 1.86 & 1.94 & 1.71 & 1.40 & 1.77 \\
Poly2 TQR                    & 1.72 & 2.04 & 1.67 & 1.74 & 1.88 & 1.91 & 1.97 & 1.78 & 1.76 & 1.64 & 1.42 & 1.58 \\
GEFCom14 Winner              & 1.29 & 1.62 & 1.17 & 1.12 & 1.25 & 1.24 & 1.24 & 1.20 & 1.26 & 1.10 & 0.82 & 1.24 \\
GEFCom14 2nd Place           & 1.24 & 1.64 & 1.23 & 1.16 & 1.25 & 1.23 & 1.25 & 1.17 & 1.31 & 1.09 & 0.84 & 1.29 \\
GEFCom14 3rd Place           & 1.73 & 1.72 & 1.46 & 1.28 & 1.29 & 1.31 & 1.32 & 1.19 & 1.37 & 1.17 & 0.86 & 1.27 \\ \hline
& \multicolumn{12}{c|}{$Q_{\mathrm{PL,B}}~[\%]$} \\ \hline
GEFCom14 Benchmark           & 3.31 & 3.88 & 3.59 & 3.61 & 4.79 & 3.57 & 4.21 & 3.99 & 4.35 & 3.77 & 3.20 & 2.85 \\ \hline
\end{tabular}}
\caption{\textcolor{review2}{Pinball-loss over all relevant tasks}}
\label{tab. explicit_pinball_loss}
\end{table}

%
%
%
\clearpage


\bibliography{biosignal}

\end{document}

%% file: nomenclature.tex
\section*{Nomenclature}

\tablehead    {\hline Symbol&Description \\ \hline}
 \tabletail    {\hline}
 \tablelasttail{\hline}
 \begin{supertabular}{|p{.15\textwidth}||p{.78\textwidth}|}
 \hline
$\bm{\mathsf{u}}[k]$ & Observation of a multivariate time series at timestep $k$ \\
$\epsilon_{\mathrm{th}}$ & Threshold defining the maximal nearest neighbors distance \\
$\hat{\boldsymbol{\theta}}$ & Estimated regression parameters \\
$\hat{\boldsymbol{\theta}}_{(q)}$ & Estimated quantile regression parameters \\
$\hat{\mathsf{G}}_{\mathrm{s}}[k]$ & Observation of the forecast surface solar radiation time series at timestep $k$ \\
$\hat{\mathsf{G}}_{\mathrm{th}}[k]$ & Observation of the forecast surface thermal radiation time series at timestep $k$ \\
$\hat{\mathsf{G}}_{\mathrm{t}}[k]$ & Observation of the forecast top net solar radiation time series at timestep $k$ \\
$\hat{\tilde{\boldsymbol{\theta}}}_{(q)}$ & Estimated parameters of a quantile regression based on the NNQF \\
$\hat{\tilde{y}}_{(q)}$ & Estimate of the $q$-quantile of $y$ given by a quantile regression based on the NNQF \\
$\hat{y}$ & Estimate of $y$ \\
$\hat{y}_{(q)}$ & Estimate of the $q$-quantile of $y$ \\
$\mathbf{X}$ & Matrix containing the training set's input vectors \\
$\mathbf{x}$ & Input vector \\
$\mathbf{x}_{i}$ & Non-nearest neighbor of $\mathbf{x}_{n}$ \\
$\mathbf{x}_{j}$ & Nearest neighbor of $\mathbf{x}_{n}$ \\
$\mathbf{x}_{n}$ & $n^{th}$ input vector in a given training/test set; with $n\in [1,N]$ \\
$\mathbf{y}$ & Vector containing the training set's desired outputs \\
$\mathbf{y}_{\mathrm{NN},n}$ & Vector containing elements $\{y_{n}; n \in J^{\mathrm{pref}}_{n}\}$ \\
$\mathsf{P}[k]$ & Observation of a photovoltaic power time series at timestep $k$ \\
$\mathsf{y}[k]$ & Observation of a time series at timestep $k$ \\
$\tilde{y}_{(q),n}$ & Empirical $q$-quantile of the elements contained in $\mathbf{y}_{\mathrm{NN},n}$ \\
$C$ & Computational effort \\
$f_{\mathrm{clock}}$ & Processor's clock rate \\
$H$ & Forecast horizon \\
$H_{1}$ & Number of lags \\
$J^{\mathrm{pref}}_{n}$ & Index set of the nearest neighbors of $\mathbf{x}_{n}$ \\
$J_{np}$ & Index set of possible nearest neighbors of $\mathbf{x}_{n}$ \\
$K$ & Length of a finite time series \\
$k$ & Timestep \\
$N$ & Number of observations in a training/test set \\
$N_{\mathrm{cores}}$ & Number of processing cores in the used computer \\
$N_{\mathrm{NN}}$ & Number of nearest neighbors \\
$p_{\mathrm{opt}}$ & $p$ index of the $J_{np}$ set containing the largest amount of elements \\
$q_{\mathrm{NN},ni}$ & Probability equal to $(i-0.5)/N_{\mathrm{NN}};i\in[1,N_{\mathrm{NN}}]$ \\
$Q_{\mathrm{PL},(q)}$ & Pinball-loss of the quantile regression with probability $q$ \\
$Q_{\mathrm{PL}}$ & Average pinball-loss across all estimated quantiles \\
$Q_{\mathrm{PL,B}}$ & \textcolor{review2}{Average pinball-loss across all estimated quantiles of the benchmark used during GEFCom14} \\
$Q_{\mathrm{R},(q)}$ & \textcolor{review2}{Percentage of values equal to or lower than the outputs of a quantile regression with probability $q$} \\
$Q_{\mathrm{Sk}}$ & \textcolor{review2}{Skill score representing the pinball loss improvement of a given approach relative to the GEFCom14 benchmark} \\
$t$ & Computation time \\
$y$ & Desired output \\
$y_{\mathrm{NN},ni}$ & $i^{th}$ value in vector $\mathbf{y}_{\mathrm{NN},n}$ \\
$y_{n}$ & $n^{th}$ desired output value in a given training/test set; with $n\in [1,N]$ \\
\end{supertabular}

%% file: main_document.bbl
\begin{thebibliography}{10}
\expandafter\ifx\csname url\endcsname\relax
  \def\url#1{\texttt{#1}}\fi
\expandafter\ifx\csname urlprefix\endcsname\relax\def\urlprefix{URL }\fi
\expandafter\ifx\csname href\endcsname\relax
  \def\href#1#2{#2} \def\path#1{#1}\fi

\bibitem{Gottwalt17}
S.~Gottwalt, J.~G{\"a}rttner, H.~Schmeck, C.~Weinhardt, Modeling and valuation
  of residential demand flexibility for renewable energy integration, IEEE
  Transactions on Smart Grid 8(6) (2017) 2565--2574.

\bibitem{Ludwig17}
N.~Ludwig, S.~Waczowicz, R.~Mikut, V.~Hagenmeyer, Mining flexibility patterns
  in energy time series from industrial processes, in: Proc., 27. Workshop
  Computational Intelligence, Dortmund, 2017, pp. 13--32.

\bibitem{Dannecker15}
L.~Dannecker, Energy Time Series Forecasting : Efficient and Accurate
  Forecasting of Evolving Time Series from the Energy Domain, 1st Edition,
  Springer Vieweg, 2015.

\bibitem{Hyndman14}
R.~J. Hyndman, G.~Athanasopoulos, Forecasting: Principles and Practice, OTexts,
  Lexington, Ky., 2014.

\bibitem{Gneiting11}
T.~Gneiting, Quantiles as optimal point forecasts, International Journal of
  Forecasting 27(2) (2011) 197--207.

\bibitem{Gneiting14}
T.~Gneiting, M.~Katzfuss, Probabilistic forecasting, Annual Review of
  Statistics and its Application 1 (2014) 125--151.

\bibitem{Fang12}
X.~Fang, S.~Misra, G.~Xue, D.~Yang, {Smart Grid} - the new and improved power
  grid: A survey, IEEE Communications Surveys Tutorials 14(4) (2012) 944--980.

\bibitem{Hong16}
T.~Hong, P.~Pinson, S.~Fan, H.~Zareipour, A.~Troccoli, R.~J. Hyndman,
  Probabilistic energy forecasting: Global energy forecasting competition 2014
  and beyond, International Journal of Forecasting 32(3) (2016) 896 -- 913.

\bibitem{GonzalezOrdiano18}
J.~{\'A}. Gonz{\'a}lez~Ordiano, S.~Waczowicz, V.~Hagenmeyer, R.~Mikut, Energy
  forecasting tools and services, Wiley Interdisciplinary Reviews: Data Mining
  and Knowledge Discovery 8~(2) (2018) e1235.

\bibitem{Fahrmeir13}
L.~Fahrmeir, T.~Kneib, S.~Lang, B.~Marx, Regression: Models, Methods and
  Applications, Springer Science \& Business Media, Berlin, 2013.

\bibitem{Bremnes04}
J.~B. Bremnes, Probabilistic wind power forecasts using local quantile
  regression, Wind Energy 7(1) (2004) 47--54.

\bibitem{Haque14}
A.~U. Haque, M.~H. Nehrir, P.~Mandal, A hybrid intelligent model for
  deterministic and quantile regression approach for probabilistic wind power
  forecasting, IEEE Transactions on Power Systems 29(4) (2014) 1663--1672.

\bibitem{Nielsen06}
H.~A. Nielsen, H.~Madsen, T.~S. Nielsen, Using quantile regression to extend an
  existing wind power forecasting system with probabilistic forecasts, Wind
  Energy 9(1-2) (2006) 95--108.

\bibitem{Gaillard16}
P.~Gaillard, Y.~Goude, R.~Nedellec, Additive models and robust aggregation for
  {GEFCom2014} probabilistic electric load and electricity price forecasting,
  International Journal of Forecasting 32(3) (2016) 1038--1050.

\bibitem{Liu17}
B.~Liu, J.~Nowotarski, T.~Hong, R.~Weron, Probabilistic load forecasting via
  quantile regression averaging on sister forecasts, IEEE Transactions on Smart
  Grid 8(2) (2017) 730--737.

\bibitem{Nagy16}
G.~I. Nagy, G.~Barta, S.~Kazi, G.~Borb{\'e}ly, G.~Simon, {GEFCom2014}:
  Probabilistic solar and wind power forecasting using a generalized additive
  tree ensemble approach, International Journal of Forecasting 32(3) (2016)
  1087--1093.

\bibitem{Zhang15}
Y.~Zhang, J.~Wang, {GEFCom2014} probabilistic solar power forecasting based on
  k-nearest neighbor and kernel density estimator, in: Proc., IEEE Power \&
  Energy Society General Meeting, IEEE, 2015, pp. 1--5.

\bibitem{Juban07}
J.~Juban, N.~Siebert, G.~N. Kariniotakis, Probabilistic short-term wind power
  forecasting for the optimal management of wind generation, in: Proc., IEEE
  Lausanne Power Tech, IEEE, 2007, pp. 683--688.

\bibitem{Xie16}
J.~Xie, T.~Hong, {GEFCom2014} probabilistic electric load forecasting: An
  integrated solution with forecast combination and residual simulation,
  International Journal of Forecasting 32(3) (2016) 1012--1016.

\bibitem{Antonanzas16}
J.~Antonanzas, N.~Osorio, R.~Escobar, R.~Urraca, F.~Martinez-de Pison,
  F.~Antonanzas-Torres, Review of photovoltaic power forecasting, Solar Energy
  136 (2016) 78--111.

\bibitem{Hong16a}
T.~Hong, S.~Fan, Probabilistic electric load forecasting: A tutorial review,
  International Journal of Forecasting.

\bibitem{Zhang14}
Y.~Zhang, J.~Wang, X.~Wang, Review on probabilistic forecasting of wind power
  generation, Renewable and Sustainable Energy Reviews 32 (2014) 255--270.

\bibitem{Cannon11}
A.~J. Cannon, Quantile regression neural networks: Implementation in {R} and
  application to precipitation downscaling, Computers \& Geosciences 37(9)
  (2011) 1277--1284.

\bibitem{Shim16}
J.~Shim, C.~Hwang, K.~Seok, et~al., Support vector quantile regression with
  varying coefficients, Computational Statistics 31~(3) (2016) 1015--1030.

\bibitem{Taylor00}
J.~W. Taylor, A quantile regression neural network approach to estimating the
  conditional density of multiperiod returns, Journal of Forecasting 19(4)
  (2000) 299--311.

\bibitem{GonzalezOrdiano16a}
J.~{\'A}. Gonz{\'a}lez~Ordiano, W.~Doneit, S.~Waczowicz, L.~Gr\"oll, R.~Mikut,
  V.~Hagenmeyer, Nearest-neighbor based non-parametric probabilistic
  forecasting with applications in photovoltaic systems, in: Proc., 26.
  Workshop Computational Intelligence, Dortmund, 2016, pp. 9--30.

\bibitem{Appino18}
R.~R. Appino, J.~{\'A}. {Gonz\'alez~Ordiano}, R.~Mikut, T.~Faulwasser,
  V.~Hagenmeyer, On the use of probabilistic forecasts in scheduling of
  renewable energy sources coupled to storages, Applied Energy 210 (2018) 1207
  -- 1218.

\bibitem{Duepmeier15}
C.~D{\"u}pmeier, K.-U. Stucky, R.~Mikut, V.~Hagenmeyer, A concept for the
  control, monitoring and visualization center in {Energy Lab 2.0}, in: Proc.,
  of the 4th D-A-CH Energy Informatics Conference, Springer International
  Publishing, Cham, 2015, pp. 83--94.

\bibitem{Hagenmeyer16}
V.~Hagenmeyer, H.~K. Cakmak, C.~D{\"u}pmeier, T.~Faulwasser, J.~Isele, H.~B.
  Keller, P.~Kohlhepp, U.~K{\"u}hnapfel, U.~Stucky, S.~Waczowicz, R.~Mikut,
  Information and communication technology in {Energy Lab 2.0}: Smart energies
  system simulation and control center with an {Open-Street-Map}-based power
  flow simulation example, Energy Technology 4 (2016) 145--162.

\bibitem{Fayyad96}
U.~Fayyad, G.~Piatetsky-Shapiro, P.~Smyth, From data mining to knowledge
  discovery in databases, AI Magazine 17 (1996) 37--54.

\bibitem{Hastie08}
T.~Hastie, R.~Tibshirani, J.~Friedman, The Elements of Statistical Learning:
  Data Mining, Inference, and Prediction, Springer, New York, 2008.

\bibitem{Stulp15}
F.~Stulp, O.~Sigaud, Many regression algorithms, one unified model: A review,
  Neural Networks 69 (2015) 60--79.

\bibitem{Koenker05}
R.~Koenker, Quantile regression, Cambridge university press, 2005.

\bibitem{Juban16}
R.~Juban, H.~Ohlsson, M.~Maasoumy, L.~Poirier, J.~Z. Kolter, A multiple
  quantile regression approach to the wind, solar, and price tracks of
  {GEFCom2014}, International Journal of Forecasting 32(3) (2016) 1094--1102.

\bibitem{Brockwell06}
P.~J. Brockwell, R.~A. Davis, Introduction to Time Series and Forecasting,
  Springer Science \& Business Media, 2006.

\bibitem{GonzalezOrdiano17}
J.~{\'A}. Gonz{\'a}lez~Ordiano, S.~Waczowicz, M.~Reischl, R.~Mikut,
  V.~Hagenmeyer, Photovoltaic power forecasting using simple data-driven models
  without weather data, Computer Science - Research and Development 32 (2017)
  237--246.

\bibitem{Hong09}
W.-C. Hong, Electric load forecasting by support vector model, Applied
  Mathematical Modelling 33(5) (2009) 2444--2454.

\bibitem{Hyndman96}
R.~J. Hyndman, Y.~Fan, Sample quantiles in statistical packages, The American
  Statistician 50(4) (1996) 361--365.

\bibitem{Langford06}
E.~Langford, Quartiles in elementary statistics, Journal of Statistics
  Education 14(3) (2006) 1--6.

\bibitem{Ma15b}
X.~Ma, X.~He, X.~Shi, A variant of k nearest neighbor quantile regression,
  Journal of Applied Statistics (2015) 1--12.

\bibitem{Huang16}
J.~Huang, M.~Perry, A semi-empirical approach using gradient boosting and
  k-nearest neighbors regression for {GEFCom2014} probabilistic solar power
  forecasting, International Journal of Forecasting 32(3) (2016) 1081--1086.

\bibitem{Gneiting07}
T.~Gneiting, F.~Balabdaoui, A.~E. Raftery, Probabilistic forecasts, calibration
  and sharpness, Journal of the Royal Statistical Society: Series B
  (Statistical Methodology) 69(2) (2007) 243--268.

\bibitem{Pinson07}
P.~Pinson, H.~A. Nielsen, J.~K. M{\o}ller, H.~Madsen, G.~N. Kariniotakis,
  Non-parametric probabilistic forecasts of wind power: Required properties and
  evaluation, Wind Energy 10(6) (2007) 497--516.

\bibitem{Mikut17}
R.~Mikut, A.~Bartschat, W.~Doneit, J.~{\'A}. {Gonz{\'a}lez~Ordiano}, B.~Schott,
  J.~Stegmaier, S.~Waczowicz, M.~Reischl, The {MATLAB} toolbox {SciXMiner}:
  User's manual and programmer's guide, Tech. rep., arXiv:1704.03298 (2017).

\bibitem{Waczowicz18_Diss}
S.~Waczowicz, {Konzept zur datengetriebenen Analyse und Modellierung des
  preisbeeinflussten Verbrauchsverhaltens}, Ph.D. thesis, Karlsruher Institut
  f\"{u}r Technologie (KIT), Fakult\"{a}t f\"{u}r Maschinenbau, submitted on
  December 22, 2017 (2018).

\bibitem{Klaiber17}
S.~Klaiber, S.~Waczowicz, I.~Konotop, D.~Westermann, R.~Mikut,
  P.~Bretschneider, {Prognose f{\"u}r preisbeeinflusstes Verbrauchsverhalten},
  at-Automatisierungstechnik 65~(3) (2017) 179--188.

\end{thebibliography}
